\documentclass[10pt, a4paper]{article}

\usepackage[final]{lrec-coling2024}
\usepackage{amsfonts}
\usepackage{amsmath}
\usepackage{booktabs}
\usepackage{color}
\usepackage{cite}
\usepackage{float} 
\usepackage{colortbl}
\usepackage{graphicx}
\usepackage{microtype}
\usepackage{multirow}
\usepackage{multicol}
\usepackage{makecell}

\newtheorem{RQ}{RQ}

\title{Does the Generator Mind its Contexts? \\
An Analysis of Generative Model Faithfulness under Context Transfer}

\name{
Xinshuo Hu$^{1}$$^{*}$\thanks{$^{*}$This work is done when Xinshuo Hu is an intern at Huawei Noah’s Ark Lab.}, 
Baotian Hu$^{1}$$^{\dagger}$\thanks{$^{\dagger}$Corresponding author.}, 
Dongfang Li$^{1}$, 
Xiaoguang Li$^{2}$, 
Lifeng Shang$^{2}$} 

\address{
$^{1}$Harbin Institute of Technology, Shenzhen, $^{2}$Huawei Noah’s Ark Lab\\
yanshek.woo@gmail.com, hubaotian@hit.edu.cn, crazyofapple@gmail.com, \\ \{lixiaoguang11, Shang.Lifeng\}@huawei.com
}

\abstract{
The present study introduces the knowledge-augmented generator, which is specifically designed to produce information that remains grounded in contextual knowledge, regardless of alterations in the context. Previous research has predominantly focused on examining hallucinations stemming from static input, such as in the domains of summarization or machine translation. However, our investigation delves into the faithfulness of generative question answering in the presence of dynamic knowledge. Our objective is to explore the existence of hallucinations arising from parametric memory when contextual knowledge undergoes changes, while also analyzing the underlying causes for their occurrence. In order to efficiently address this issue, we propose a straightforward yet effective measure for detecting such hallucinations. Intriguingly, our investigation uncovers that all models exhibit a tendency to generate previous answers as hallucinations. To gain deeper insights into the underlying causes of this phenomenon, we conduct a series of experiments that verify the critical role played by context in hallucination, both during training and testing, from various perspectives.
 \\ \newline  }

\begin{document}

\maketitleabstract

\section{Introduction}
\label{Sec:Tntroduction}

\begin{figure}[t]
  \centering
  \includegraphics[width=0.48\textwidth]{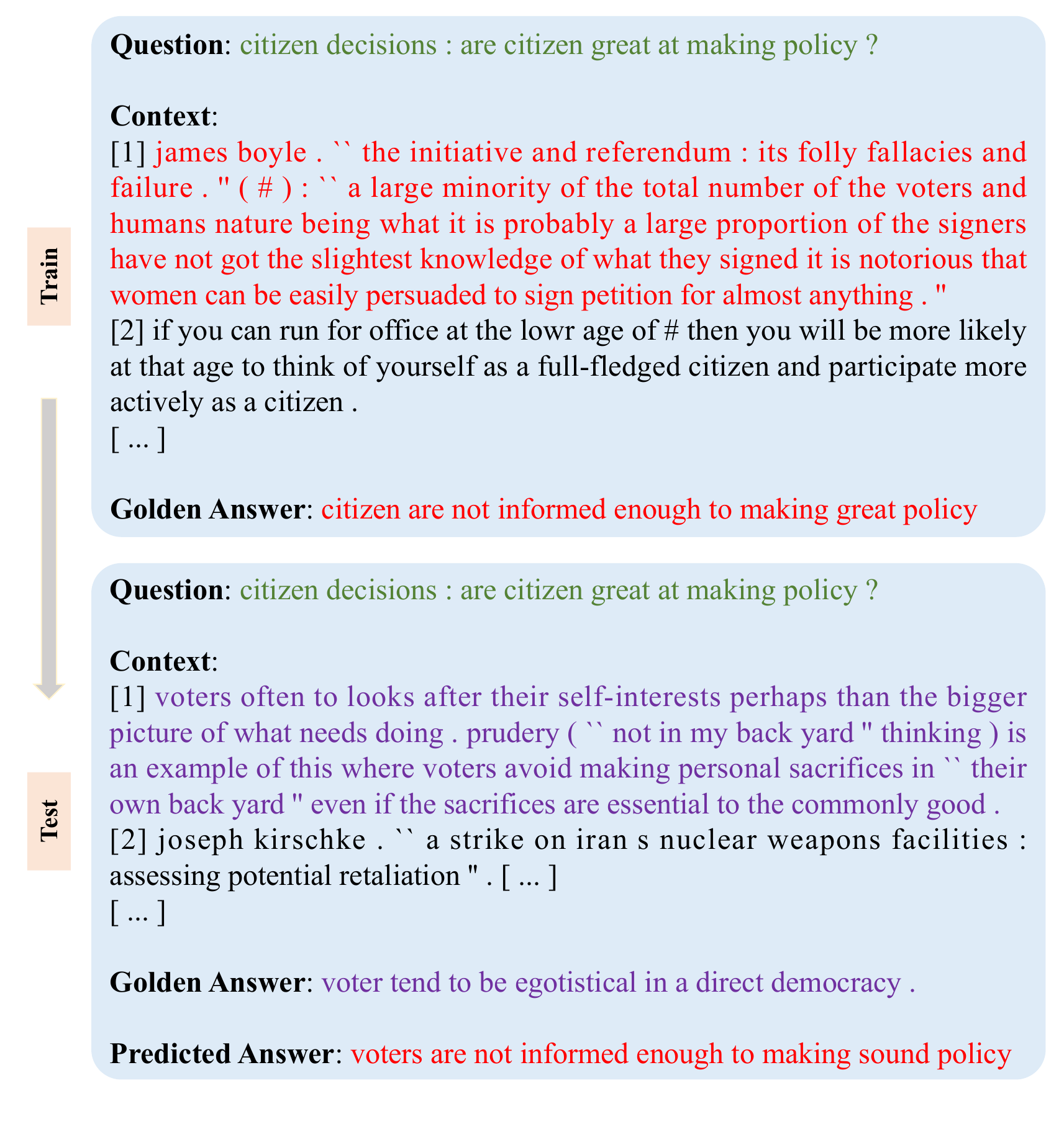}
  \caption{An example of generated hallucination from training memory. The model disregards \textcolor[RGB]{112,48,160}{the transferred contextual knowledge} and predicts \textcolor[RGB]{255,0,0}{an out-of-date answer} that was present in its original training data when answering \textcolor[RGB]{84,130,53}{the same question}. Non-essential details are ignored by [...].}
  \label{Fig:Introduction_Case}
\end{figure}

Knowledge-augmented text generation method (e.g. RAG~\citep{DBLP:conf/nips/LewisPPPKGKLYR020}, FiD~\citep{DBLP:conf/eacl/IzacardG21}), and Atlas~\citep{DBLP:journals/corr/abs-2208-03299}, have demonstrated state-of-the-art (SOTA) performance across various NLP tasks. The paradigm of generating text using external knowledge offers the advantage of plug-and-play through non-parametric contextual knowledge. In contrast, parametric knowledge embedded within models necessitates retraining for updates~\citep{DBLP:journals/corr/abs-2202-01110}. A faithful knowledge-augmented generator should consistently produce output that aligns with the contextual grounding~\citep{DBLP:journals/corr/abs-2202-03629}. However, the presence of hallucinations originating from parametric memory (see \autoref{Fig:Introduction_Case}) poses a significant challenge for practical text generation applications~\citep{DBLP:conf/acl/MaynezNBM20,DBLP:conf/acl/ZhangMTML20}.

The investigation of the faithfulness of generative models in the presence of dynamic contextual knowledge remains an ongoing research area. Previous studies have primarily focused on analyzing hallucinations in scenarios where the input texts during training and testing are independent, such as in summarization\citep{DBLP:conf/naacl/PagnoniBT21,DBLP:conf/acl/LadhakD0CM22,DBLP:journals/corr/abs-2205-12854} or machine translation\citep{DBLP:conf/naacl/RaunakMJ21,DBLP:conf/amta/MullerRS20}. While knowledge-dynamic question answering has garnered attention in several works~\citep{DBLP:conf/emnlp/MinMHZ20,DBLP:conf/emnlp/LongprePCRD021,DBLP:conf/emnlp/ZhangC21,DBLP:conf/nips/ChenWWW21,DBLP:conf/sigir/WangJY22,DBLP:conf/icml/LiskaKGTSAdSZYG22,DBLP:journals/corr/abs-2207-13332,DBLP:journals/corr/abs-2311-09149}, only a few studies have systematically quantified the extent of model faithfulness or analyzed the circumstances and reasons behind hallucination generation in the presence of dynamic contextual knowledge~\citep{DBLP:conf/emnlp/LongprePCRD021,DBLP:conf/acl/WestQGC22}. 
In this study, we define \textit{context transfer} as the process of contextual knowledge changing while the question remains the same. Specifically, the generative model is trained on old knowledge but evaluated on new knowledge instances. Our analysis focuses on \textit{memory hallucination} which refers to hallucinations generated by parametric knowledge during context transfer. 

In this work, our objective is to assess the faithfulness of generative models in the context of context transfer, focusing on two primary research questions:
\begin{RQ}
     To what extent does the generative model exhibit faithfulness under context transfer?
     \label{rq:1}
\end{RQ}
\begin{RQ}
    What are the underlying reasons for the occurrence of memory hallucination?
    \label{rq:2}
\end{RQ}
To address these research questions, we first define the context transfer task and introduce a novel metric for measuring hallucination (\S\ref{Sec:Method}). Subsequently, we conduct comprehensive experiments involving multiple models to investigate Research Question~\ref{rq:1}. Our findings indicate that models do not consistently exhibit grounded behavior in the presence of context transfer (\S\ref{Sec:Results}). To gain deeper insights into the issue raised in Research Question~\ref{rq:2}, we perform an in-depth analysis of contextual knowledge, revealing that the presence of noisy and irrelevant contexts hinders models from effectively capturing the desired question-context-answer correlation (\S\ref{Sec:Analysis}). 

\section{Related Work}

\subsection{Faithful Natural Language Generation}

Faithful natural language generation (NLG) aims to generate text that is both faithful and consistent with the input information, while avoiding hallucination \citep{DBLP:journals/corr/abs-2203-05227,DBLP:journals/corr/abs-2202-03629}. 
In recent years, there has been a growing interest in understanding factual errors in summarization \citep{DBLP:conf/naacl/PagnoniBT21,DBLP:conf/acl/LadhakD0CM22,DBLP:journals/corr/abs-2205-12854} and machine translation \citep{DBLP:conf/amta/MullerRS20,DBLP:conf/naacl/RaunakMJ21}.
Additionally, there have been studies focusing on knowledge faithfulness in question answering \citep{DBLP:conf/naacl/KrishnaRI21,DBLP:journals/corr/abs-2112-13432,DBLP:conf/emnlp/LongprePCRD021} and dialogue response generation \citep{DBLP:conf/emnlp/HonovichCANSA21,DBLP:conf/naacl/DziriMYZR22}. For more details, we refer readers to the surveys \citep{DBLP:journals/corr/abs-2203-05227,DBLP:journals/corr/abs-2202-03629}. Although factoid hallucination has been extensively studied, our work focuses on a broader scope by considering non-factoid information, such as debates and opinions.

\subsection{Context Transfer}

Context transfer in NLG involves models adapting to dynamically provided information rather than relying solely on pre-learned parameters. This aspect has been explored in studies on Wikipedia writing by \citet{DBLP:conf/naacl/PrabhumoyeQG19} and \citet{DBLP:conf/acl/WestQGC22}, investigating the model grounding ability. Furthermore, several works have addressed question answering in the context of dynamic knowledge \citep{DBLP:conf/emnlp/MinMHZ20,DBLP:conf/emnlp/LongprePCRD021,DBLP:conf/emnlp/ZhangC21,DBLP:conf/nips/ChenWWW21,DBLP:conf/sigir/WangJY22,DBLP:conf/icml/LiskaKGTSAdSZYG22,DBLP:journals/corr/abs-2207-13332}. The most similar work is \citet{DBLP:conf/emnlp/LongprePCRD021}, which focused on entity-based knowledge conflict and was under the open-domain setting. However, we investigate long-form question answering (LFQA), where we transfer the entire knowledge text rather than solely editing entities. All transferred knowledge remains relevant and aligned with the real world, as false contextual information may conflict with pre-learned knowledge and potentially induce hallucinations in the model.
\section{Methods}
\label{Sec:Method}

\subsection{Task: Question Answering under Context Transfer}
\label{Sec:Method:Task}

Context transfer necessitates the model's ability to generate a novel answer based on newly acquired knowledge for the same question during training.
To begin, we employ a dataset $D$ consisting of two partitions, namely $D_{train}$ and $D_{test}$. Our initial step involves training a knowledge-grounded generative model on the training examples $(q_i, c_i, a_i) \in D_{train}$, where $q_i$ represents the question,
$c_i$ consists of contextual sentences comprising positive ($c_i^+$) and negative ($c_i^-$) contextual knowledge, and $a_i$ denotes the golden reference answer. Subsequently, the model is evaluated using examples $(q_j, \hat{c_j}) \in D_{test}$, wherein the query $q_j$ can be found in $D_{train}$, while the contextual knowledge $c_j$ is transferred to $\hat{c_j}$. 

Our primary focus lies in abstractive long-form question answering. We consider entity-based question answering to be straightforward, as hallucination can be mitigated or even resolved through extraction-augmentation and post-editing techniques. To construct a relevant benchmark, we utilize query-based summarization data from Debatepedia~\citep{DBLP:conf/acl/NemaKLR17}, primarily due to its highly abstract nature and natural conditions for context transfer.
In contrast to previous research~\citep{DBLP:conf/emnlp/LongprePCRD021}, we adopt a more natural setting where the transferred contextual knowledge is factual as well. Furthermore, we ensure that the questions are answerable, considering it a necessary requirement. This precaution is taken because we have observed that models tend to generate hallucinatory responses when the contextual knowledge does not contribute to answering the question effectively.

\subsection{Measure: Margin Failure Rate}
\label{Sec:Method:Measure}

As illustrated in \autoref{Fig:Introduction_Case}, the trained model exhibits a failure in grounding transferred contextual knowledge, resulting in the generation of answers that are not properly aligned with the given contexts. This phenomenon is referred to as a \textit{grounding failure of context transfer}. 

To determine whether a predicted answer $\hat{a}$ represents a grounding failure of context transfer, we introduce the concept of \textit{margin grounding failure} ($\mathcal{MF}$) as follows:
\begin{equation}
\begin{aligned}
\mathcal{MF}(\Phi) = \left\{
  \begin{aligned}
    & 1, \Phi (\hat{a}, r_{train}) > m \cdot \Phi (\hat{a}, r_{test}) \\ 
    & 0, \Phi (\hat{a}, r_{train}) \le m \cdot \Phi (\hat{a}, r_{test})
  \end{aligned}
\right.
\end{aligned}
\end{equation}
\label{eq:MGF}
where $m$ represents the hyperparameter margin, and $\Phi$ is a basic metric (e.g. ROUGE) to measure the similarity between the predicted answer $\hat{a}$ and golden reference $r$. The reference $r$ comes from either the train or test set ($r_{train}$ from the train set or $r_{test}$ from the test set), which can be the golden answer or the contextual knowledge\footnote{In cases where there are multiple references, individual scores are calculated, and the maximum score is selected.}. 

It is important to note that grounding failure is a binary label assigned to each case. To statistically probe the faithfulness over the test set, we propose to measure the percentage of grounding failure of context transfer. So the \textit{margin failure rate} ($\mathcal{MFR}$) is defined as:
\begin{equation}
\begin{aligned}
\mathcal{MFR}(\Phi) =  \frac{1}{N} \sum_{i=1}^{N} \mathcal{MF}_{i}(\Phi) .
\end{aligned}
\end{equation}
\label{eq:MFR}
In this work, we use BERT-SCORE~\citep{DBLP:conf/iclr/ZhangKWWA20} as our basic metric $\Phi$. For our experiments, we set the margin $ m$ to a value of $1.25$ based on intuition, which has a relatively strong correlation with Pearson Correlation of $0.43$ with human evaluation on our development set.

\section{Results}
\label{Sec:Results}

\begin{table}
  \centering
  \small
  \begin{tabular}{lcc}
    \toprule
    \multirow{2}{*}{\bfseries Model} & \multicolumn{2}{c}{\bfseries Decoding Strategy} \\
    \cmidrule(r){2-3} 
       & {\bfseries Greedy} & {\bfseries Beam Search}  \\
    \hline

    T5$_{small}$             & 7.69 & 8.19 \\
    T5$_{base}$              & 7.53 & 6.19 \\
    BART$_{base}$           & 9.20 & 10.87 \\
    BART$_{large}$          & 7.86 & 8.36 \\
    BART$_{large-xsum}$      & 8.03 & 7.19 \\
    \hline
    FiD (T5$_{small}$)       & 11.37 & 9.53 \\
    FiD (T5$_{base}$)        & 11.04 & 10.03 \\
    FiD (BART$_{base}$)      & 13.88 & 12.71 \\
    FiD (BART$_{large}$)     & 10.03 & 8.86 \\
    FiD (BART$_{large-xsum}$)& 15.38 & 14.55 \\

    \bottomrule
  \end{tabular}
  
  \caption{The $\mathcal{MFR}$(BERT-Score) results of different models. We generate text by greedy and beam search (beam=4) decoding strategy.}
  \label{Tab:Model_comparison}
\end{table}

In this study, we present the outcomes obtained from two prominent state-of-the-art sequence-to-sequence (seq2seq) pre-trained models, namely BART~\citep{DBLP:conf/acl/LewisLGGMLSZ20} and T5~\citep{DBLP:journals/jmlr/RaffelSRLNMZLL20}, in the context of question answering (QA) tasks. Besides the vanilla transformer architecture, we also incorporate the FiD method ~\citep{DBLP:conf/eacl/IzacardG21} owing to its efficient and effective utilization of extensive document collections.
The model selection process is based on the ROUGE-L score achieved on the development set.

\paragraph{All models have memory hallucination under context transfer.} 
The $\mathcal{MFR}$(BERT-Score) results of various models under context transfer are presented in \autoref{Tab:Model_comparison}. It is observed that all the models exhibit the phenomenon of memory hallucination during context transfer, albeit to varying degrees. The choice of decoding strategies does not appear to have a significant impact on the generation of hallucinations. Specifically, the FiD method demonstrates a higher occurrence of context transfer grounding failure compared to the vanilla transformer. This can be attributed to the fact that FiD has a tendency to memorize the question-answer pairs, as the questions are duplicated for each context.

\section{Analysis}
\label{Sec:Analysis}

	

\begin{figure}[t]
  \centering
  \includegraphics[width=0.49\textwidth]{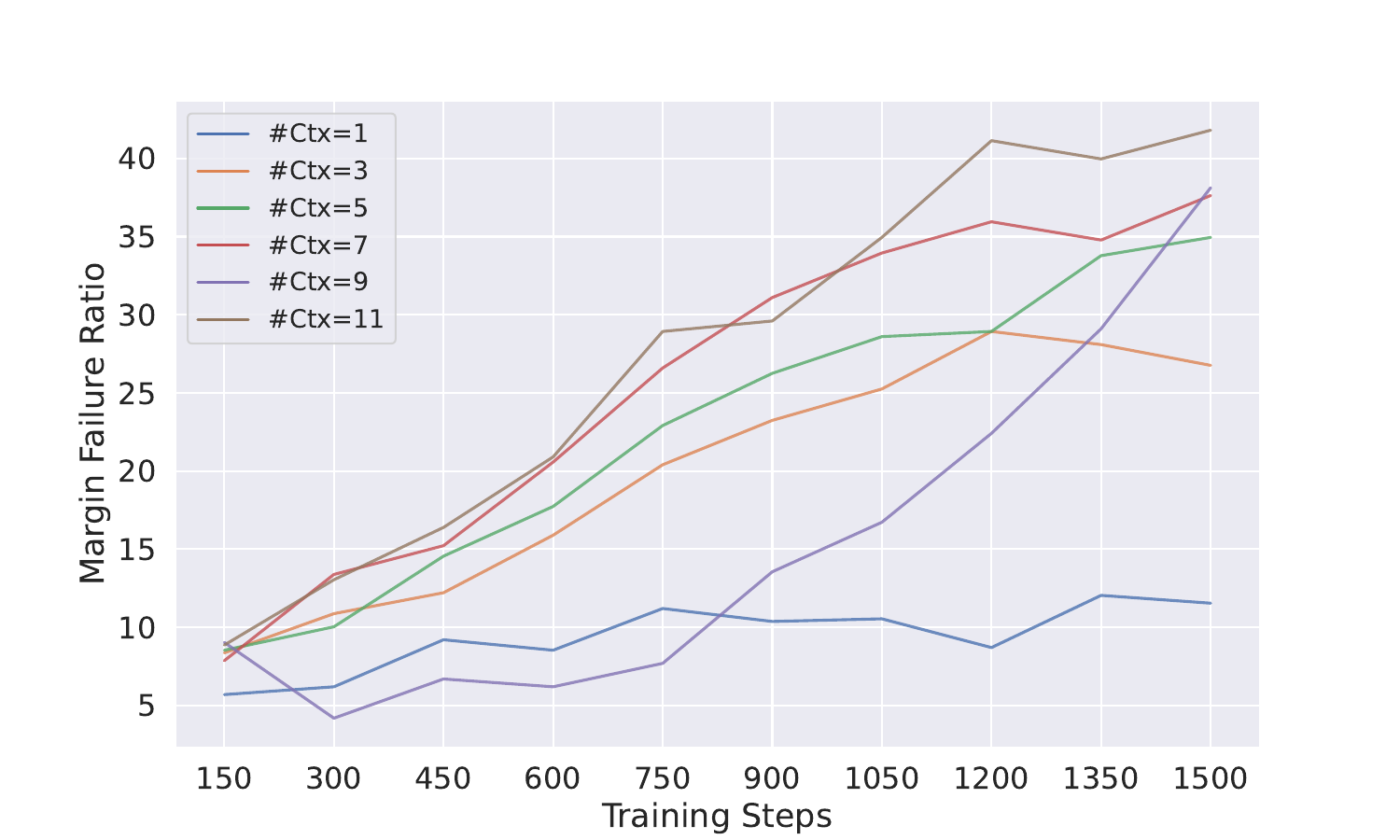}
  \caption{The influence of the scale of contextual knowledge and training step on $\mathcal{MFR}$(BERT-Score).}
  \label{Fig:Ctx_scale_impact}
\end{figure}

In this section, we endeavor to elucidate the intricate interplay between causality and its impact on model faithfulness within the realm of context transfer. To this end, we embark upon a series of rigorous experiments, wherein we manipulate contextual factors from various perspectives, in order to derive meaningful insights.
We conduct all the analysis on FiD(BART$_{large-xsum}$).

\begin{figure}[t]
  \centering
  \includegraphics[width=0.46\textwidth]{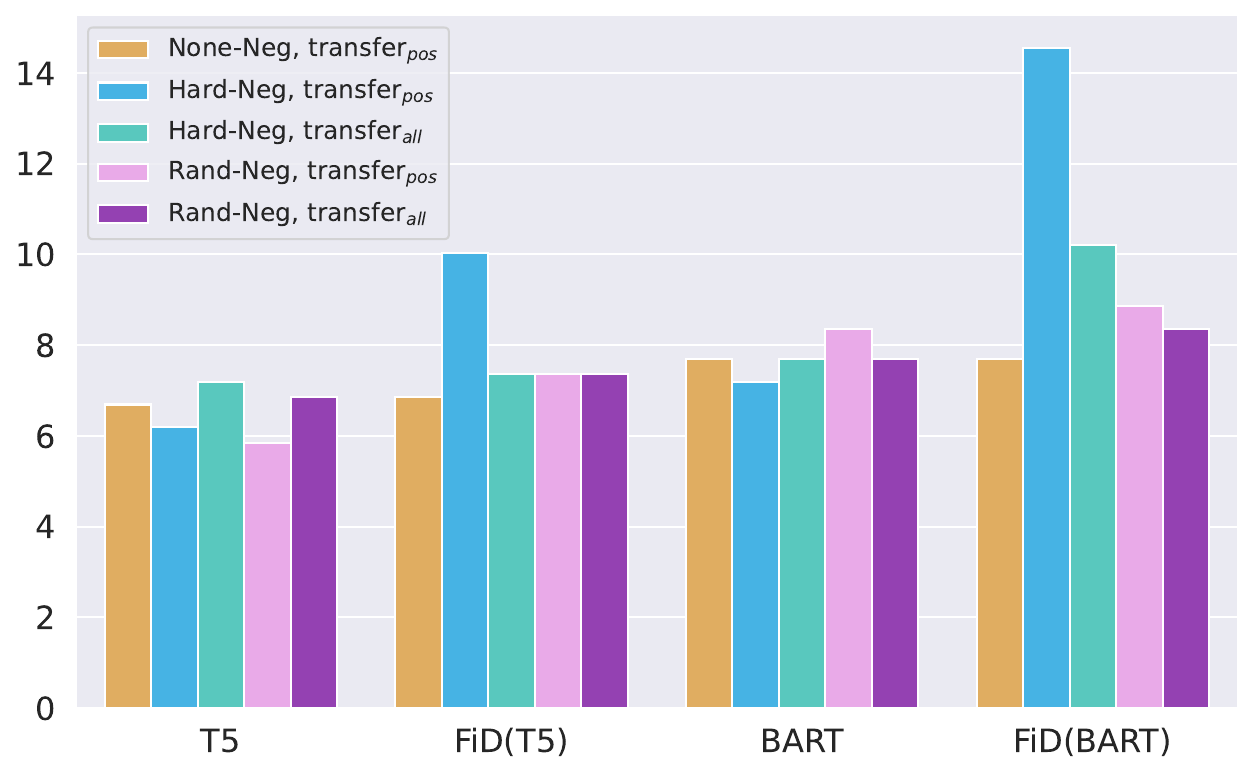}
  \caption{The $\mathcal{MFR}$(BERT-Score) results over different settings of contexts.}
  \label{Fig:neg_ctx_impact}
\end{figure}

\paragraph{Impact of Contextual Knowledge Scale} We examine the effect of varying the scale of contextual knowledge on the performance of FiD (BART$_{large-xsum}$) as measured by the $\mathcal{MFR}$(BERT-Score). It becomes evident that the $\mathcal{MFR}$ value increases proportionally with the expansion of the context scale (\autoref{Fig:Ctx_scale_impact}). This surplus of noisy contexts hampers the model's ability to ground itself in accurate knowledge and introduces confusion during the generation process, as elaborated upon later in \autoref{Fig:neg_ctx_impact}. 
Therefore, it becomes crucial to strike a balance between the quantity of information retrieved and the presence of noise, particularly in practical applications where obtaining more knowledge through an imperfect retriever holds significance.
Furthermore, it is worth noting that training the model for an extended duration may lead to overfitting on question-answer spurious correlations. 
Notably, the $\mathcal{MFR}$(BERT-Score) can reach as high as $20$ after a mere $600$ training steps, equivalent to approximately four epochs.

\paragraph{Impact of Irrelevant Noisy Context} The presence of irrelevant noisy context can have a detrimental effect on faithful generation during both the training and testing phases. In our experiments, we explore different settings of contextual knowledge using the T5$_{base}$ and BART$_{large-xsum}$. During the training process, we introduce negative contexts using two different methods: retrieval-based methods (referred to as Hard-Neg) nd random sampling (referred to as Rand-Neg). For testing, we consider two scenarios: transferring only the positive context while keeping the negative contexts unchanged (referred to as transfer$_{pos}$), or transferring both the positive and negative contexts by replacing the latter with random ones (referred to as transfer$_{all}$). The detailed settings are as follows:
\begin{itemize}
\item [1)] 
\noindent\textbf{None Negative Contexts (None-Neg):} Only positive contextual knowledge is provided during training. During testing, we transfer only the positive knowledge (\textbf{transfer$_{pos}$}).
\item [2)]
\noindent\textbf{Hard Negative Contexts (Hard-Neg):} In this setting, we provide the positive contextual knowledge along with retrieved hard negative knowledge using BM25. This setting is more realistic as it involves retrieving external knowledge in an open domain. During testing, \textbf{transfer$_{pos}$} refers to transferring only the positive knowledge, while \textbf{transfer$_{all}$} refers to transferring both the positive and negative knowledge, with the negative knowledge being randomly sampled.
\item [3)]
\noindent\textbf{Random Negative Contexts (Rand-Neg):} Similar to the Hard-Neg setting, we provide the positive contextual knowledge, but pair it with randomly sampled negative knowledge. The testing scenarios (\textbf{transfer${pos}$} and \textbf{transfer$_{all}$}) remain the same as in the Hard-Neg setting.
\end{itemize}

The final comparative results are presented in ~\autoref{Fig:neg_ctx_impact}. Notably, there is a drop on $\mathcal{MFR}$(BERT-Score) for the FiD architecture when tested on transfer$_{all}$, specially trained on hard negative contexts. The presence of hard negative contexts poses a challenging confounding factor, as it may induce models to learn spurious correlations, given that retrieved knowledge is often more relevant to the question than sampled knowledge. Furthermore, our findings align with the conclusions drawn from \autoref{Fig:Ctx_scale_impact}, indicating that the inclusion of negative contexts significantly increases the occurrence of margin grounding failure. However, it is worth noting that the vanilla transformer architecture exhibits robustness against negative contexts, displaying insensitivity to contextual disturbance. Upon comparing transfer$_{pos}$ with transfer$_{all}$, we observe that the model unintentionally grounds its answers on irrelevant knowledge when negative contexts are transferred, leading to unexpected changes in the generated answers.



\section{Conclusion}

This study endeavors to explore the phenomenon of memory hallucination in the realm of context transfer. Our investigation entails the comprehensive examination of multiple models, unveiling potential deficiencies in their ability to faithfully align contextual knowledge. Furthermore, our research emphasizes the pivotal role played by context in the manifestation of hallucinations during both training and testing phases. Despite the apparent rarity of memory hallucination, it represents a critical concern that demands attention for the attainment of veracious natural language generation in practical settings. We anticipate that this research will contribute to a more profound comprehension of the faithfulness of generative models.
\section*{Limitations}
\label{sec:appendix:Limitation_and_Future Work}
\paragraph{Benchmark Dataset} Acquiring suitable datasets for long-form abstractive Question Answering (QA) in the context of context transfer poses a significant challenge. Although Debatepedia may initially seem appropriate for such experiments, the reliability of its data scale and quality is questionable, thereby limiting our ability to investigate the factors that influence answer faithfulness.
We anticipate that future research will explore additional domains and levels of context transfer, expanding the scope of investigation.

\paragraph{Evaluation Metrics} Existing automatic evaluation metrics Existing automatic evaluation metrics demonstrate limited correlation with human evaluations. Therefore, it is crucial to propose an alternative methodology for systematically assessing large-scale results, with the aim of reducing the variance inherent in small-scale data.

\paragraph{Evaluation Models} Owing to constraints in resources, comprehensive experimentation on the prevalent large language models, has not been undertaken. Nonetheless, we have intentions to incorporate experiments pertaining to large language models in our future endeavors, contingent upon the feasibility thereof.

\paragraph{Faithfulness Improvement} The primary goal of faithfulness probing is to establish a generative model that faithfully incorporates and aligns with the provided context. Nevertheless, this work lacks methodologies to enhance the faithfulness of generative models. Consequently, we try to advance this investigation by exploring the causal factors behind hallucination and proposing viable solutions to address this intricate challenge.

\section*{Acknowledgments}
We thank Shaobo Li and Jimmy Wu for their insightful suggestions and invaluable feedback. This work is supported by grants: Natural
Science Foundation of China (No. 62376067).

\nocite{*}
\section*{References}\label{sec:reference}
\bibliographystyle{lrec-coling2024-natbib}
\bibliography{lrec-coling2024}

\begin{thebibliography}{40}
\expandafter\ifx\csname natexlab\endcsname\relax\def\natexlab#1{#1}\fi

\bibitem[{Chen et~al.(2021)Chen, Wang, and Wang}]{DBLP:conf/nips/ChenWWW21}
Wenhu Chen, Xinyi Wang, and William~Yang Wang. 2021.
\newblock \href {https://datasets-benchmarks-proceedings.neurips.cc/paper/2021/hash/1f0e3dad99908345f7439f8ffabdffc4-Abstract-round2.html} {A dataset for answering time-sensitive questions}.
\newblock In \emph{Proceedings of the Neural Information Processing Systems Track on Datasets and Benchmarks 1, NeurIPS Datasets and Benchmarks 2021, December 2021, virtual}.

\bibitem[{Chen et~al.(2023)Chen, Li, Zhao, Hu, and Zhang}]{DBLP:journals/corr/abs-2311-09149}
Ziyang Chen, Dongfang Li, Xiang Zhao, Baotian Hu, and Min Zhang. 2023.
\newblock \href {https://doi.org/10.48550/ARXIV.2311.09149} {Temporal knowledge question answering via abstract reasoning induction}.
\newblock \emph{CoRR}, abs/2311.09149.

\bibitem[{Dreyer et~al.(2021)Dreyer, Liu, Nan, Atluri, and Ravi}]{DBLP:journals/corr/abs-2108-02859}
Markus Dreyer, Mengwen Liu, Feng Nan, Sandeep Atluri, and Sujith Ravi. 2021.
\newblock \href {http://arxiv.org/abs/2108.02859} {Analyzing the abstractiveness-factuality tradeoff with nonlinear abstractiveness constraints}.
\newblock \emph{CoRR}, abs/2108.02859.

\bibitem[{Dziri et~al.(2022)Dziri, Milton, Yu, Za{\"{\i}}ane, and Reddy}]{DBLP:conf/naacl/DziriMYZR22}
Nouha Dziri, Sivan Milton, Mo~Yu, Osmar~R. Za{\"{\i}}ane, and Siva Reddy. 2022.
\newblock \href {https://doi.org/10.18653/v1/2022.naacl-main.387} {On the origin of hallucinations in conversational models: Is it the datasets or the models?}
\newblock In \emph{Proceedings of the 2022 Conference of the North American Chapter of the Association for Computational Linguistics: Human Language Technologies, {NAACL} 2022, Seattle, WA, United States, July 10-15, 2022}, pages 5271--5285. Association for Computational Linguistics.

\bibitem[{Goyal et~al.(2022)Goyal, Xu, Li, and Durrett}]{DBLP:conf/acl/GoyalXLD22}
Tanya Goyal, Jiacheng Xu, Junyi~Jessy Li, and Greg Durrett. 2022.
\newblock \href {https://doi.org/10.18653/v1/2022.findings-acl.163} {Training dynamics for text summarization models}.
\newblock In \emph{Findings of the Association for Computational Linguistics: {ACL} 2022, Dublin, Ireland, May 22-27, 2022}, pages 2061--2073. Association for Computational Linguistics.

\bibitem[{Grusky et~al.(2018)Grusky, Naaman, and Artzi}]{DBLP:conf/naacl/GruskyNA18}
Max Grusky, Mor Naaman, and Yoav Artzi. 2018.
\newblock \href {https://doi.org/10.18653/v1/n18-1065} {Newsroom: {A} dataset of 1.3 million summaries with diverse extractive strategies}.
\newblock In \emph{Proceedings of the 2018 Conference of the North American Chapter of the Association for Computational Linguistics: Human Language Technologies, {NAACL-HLT} 2018, New Orleans, Louisiana, USA, June 1-6, 2018, Volume 1 (Long Papers)}, pages 708--719. Association for Computational Linguistics.

\bibitem[{Honovich et~al.(2021)Honovich, Choshen, Aharoni, Neeman, Szpektor, and Abend}]{DBLP:conf/emnlp/HonovichCANSA21}
Or~Honovich, Leshem Choshen, Roee Aharoni, Ella Neeman, Idan Szpektor, and Omri Abend. 2021.
\newblock \href {https://doi.org/10.18653/v1/2021.emnlp-main.619} {{\textdollar}q{\^{}}2{\textdollar}: Evaluating factual consistency in knowledge-grounded dialogues via question generation and question answering}.
\newblock In \emph{Proceedings of the 2021 Conference on Empirical Methods in Natural Language Processing, {EMNLP} 2021, Virtual Event / Punta Cana, Dominican Republic, 7-11 November, 2021}, pages 7856--7870. Association for Computational Linguistics.

\bibitem[{Hu et~al.(2015)Hu, Chen, and Zhu}]{DBLP:conf/emnlp/HuCZ15}
Baotian Hu, Qingcai Chen, and Fangze Zhu. 2015.
\newblock \href {https://doi.org/10.18653/V1/D15-1229} {{LCSTS:} {A} large scale chinese short text summarization dataset}.
\newblock In \emph{Proceedings of the 2015 Conference on Empirical Methods in Natural Language Processing, {EMNLP} 2015, Lisbon, Portugal, September 17-21, 2015}, pages 1967--1972. The Association for Computational Linguistics.

\bibitem[{Izacard and Grave(2021)}]{DBLP:conf/eacl/IzacardG21}
Gautier Izacard and Edouard Grave. 2021.
\newblock \href {https://doi.org/10.18653/v1/2021.eacl-main.74} {Leveraging passage retrieval with generative models for open domain question answering}.
\newblock In \emph{Proceedings of the 16th Conference of the European Chapter of the Association for Computational Linguistics: Main Volume, {EACL} 2021, Online, April 19 - 23, 2021}, pages 874--880. Association for Computational Linguistics.

\bibitem[{Izacard et~al.(2022)Izacard, Lewis, Lomeli, Hosseini, Petroni, Schick, Dwivedi{-}Yu, Joulin, Riedel, and Grave}]{DBLP:journals/corr/abs-2208-03299}
Gautier Izacard, Patrick Lewis, Maria Lomeli, Lucas Hosseini, Fabio Petroni, Timo Schick, Jane Dwivedi{-}Yu, Armand Joulin, Sebastian Riedel, and Edouard Grave. 2022.
\newblock \href {https://doi.org/10.48550/arXiv.2208.03299} {Few-shot learning with retrieval augmented language models}.
\newblock \emph{CoRR}, abs/2208.03299.

\bibitem[{Ji et~al.(2022)Ji, Lee, Frieske, Yu, Su, Xu, Ishii, Bang, Madotto, and Fung}]{DBLP:journals/corr/abs-2202-03629}
Ziwei Ji, Nayeon Lee, Rita Frieske, Tiezheng Yu, Dan Su, Yan Xu, Etsuko Ishii, Yejin Bang, Andrea Madotto, and Pascale Fung. 2022.
\newblock \href {http://arxiv.org/abs/2202.03629} {Survey of hallucination in natural language generation}.
\newblock \emph{CoRR}, abs/2202.03629.

\bibitem[{Kasai et~al.(2022)Kasai, Sakaguchi, Takahashi, Bras, Asai, Yu, Radev, Smith, Choi, and Inui}]{DBLP:journals/corr/abs-2207-13332}
Jungo Kasai, Keisuke Sakaguchi, Yoichi Takahashi, Ronan~Le Bras, Akari Asai, Xinyan Yu, Dragomir~R. Radev, Noah~A. Smith, Yejin Choi, and Kentaro Inui. 2022.
\newblock \href {https://doi.org/10.48550/arXiv.2207.13332} {Realtime {QA:} what's the answer right now?}
\newblock \emph{CoRR}, abs/2207.13332.

\bibitem[{Kingma and Ba(2015)}]{DBLP:journals/corr/KingmaB14}
Diederik~P. Kingma and Jimmy Ba. 2015.
\newblock \href {http://arxiv.org/abs/1412.6980} {Adam: {A} method for stochastic optimization}.
\newblock In \emph{3rd International Conference on Learning Representations, {ICLR} 2015, San Diego, CA, USA, May 7-9, 2015, Conference Track Proceedings}.

\bibitem[{Krishna et~al.(2021)Krishna, Roy, and Iyyer}]{DBLP:conf/naacl/KrishnaRI21}
Kalpesh Krishna, Aurko Roy, and Mohit Iyyer. 2021.
\newblock \href {https://doi.org/10.18653/v1/2021.naacl-main.393} {Hurdles to progress in long-form question answering}.
\newblock In \emph{Proceedings of the 2021 Conference of the North American Chapter of the Association for Computational Linguistics: Human Language Technologies, {NAACL-HLT} 2021, Online, June 6-11, 2021}, pages 4940--4957. Association for Computational Linguistics.

\bibitem[{Ladhak et~al.(2022)Ladhak, Durmus, He, Cardie, and McKeown}]{DBLP:conf/acl/LadhakD0CM22}
Faisal Ladhak, Esin Durmus, He~He, Claire Cardie, and Kathleen~R. McKeown. 2022.
\newblock \href {https://doi.org/10.18653/v1/2022.acl-long.100} {Faithful or extractive? on mitigating the faithfulness-abstractiveness trade-off in abstractive summarization}.
\newblock In \emph{Proceedings of the 60th Annual Meeting of the Association for Computational Linguistics (Volume 1: Long Papers), {ACL} 2022, Dublin, Ireland, May 22-27, 2022}, pages 1410--1421. Association for Computational Linguistics.

\bibitem[{Lewis et~al.(2020{\natexlab{a}})Lewis, Liu, Goyal, Ghazvininejad, Mohamed, Levy, Stoyanov, and Zettlemoyer}]{DBLP:conf/acl/LewisLGGMLSZ20}
Mike Lewis, Yinhan Liu, Naman Goyal, Marjan Ghazvininejad, Abdelrahman Mohamed, Omer Levy, Veselin Stoyanov, and Luke Zettlemoyer. 2020{\natexlab{a}}.
\newblock \href {https://doi.org/10.18653/v1/2020.acl-main.703} {{BART:} denoising sequence-to-sequence pre-training for natural language generation, translation, and comprehension}.
\newblock In \emph{Proceedings of the 58th Annual Meeting of the Association for Computational Linguistics, {ACL} 2020, Online, July 5-10, 2020}, pages 7871--7880. Association for Computational Linguistics.

\bibitem[{Lewis et~al.(2020{\natexlab{b}})Lewis, Perez, Piktus, Petroni, Karpukhin, Goyal, K{\"{u}}ttler, Lewis, Yih, Rockt{\"{a}}schel, Riedel, and Kiela}]{DBLP:conf/nips/LewisPPPKGKLYR020}
Patrick S.~H. Lewis, Ethan Perez, Aleksandra Piktus, Fabio Petroni, Vladimir Karpukhin, Naman Goyal, Heinrich K{\"{u}}ttler, Mike Lewis, Wen{-}tau Yih, Tim Rockt{\"{a}}schel, Sebastian Riedel, and Douwe Kiela. 2020{\natexlab{b}}.
\newblock \href {https://proceedings.neurips.cc/paper/2020/hash/6b493230205f780e1bc26945df7481e5-Abstract.html} {Retrieval-augmented generation for knowledge-intensive {NLP} tasks}.
\newblock In \emph{Advances in Neural Information Processing Systems 33: Annual Conference on Neural Information Processing Systems 2020, NeurIPS 2020, December 6-12, 2020, virtual}.

\bibitem[{Li et~al.(2022{\natexlab{a}})Li, Su, Cai, Wang, and Liu}]{DBLP:journals/corr/abs-2202-01110}
Huayang Li, Yixuan Su, Deng Cai, Yan Wang, and Lemao Liu. 2022{\natexlab{a}}.
\newblock \href {http://arxiv.org/abs/2202.01110} {A survey on retrieval-augmented text generation}.
\newblock \emph{CoRR}, abs/2202.01110.

\bibitem[{Li et~al.(2022{\natexlab{b}})Li, Wu, Chen, Liu, Xiao, and Wu}]{DBLP:journals/corr/abs-2203-05227}
Wei Li, Wenhao Wu, Moye Chen, Jiachen Liu, Xinyan Xiao, and Hua Wu. 2022{\natexlab{b}}.
\newblock \href {https://doi.org/10.48550/arXiv.2203.05227} {Faithfulness in natural language generation: {A} systematic survey of analysis, evaluation and optimization methods}.
\newblock \emph{CoRR}, abs/2203.05227.

\bibitem[{Liska et~al.(2022)Liska, Kocisk{\'{y}}, Gribovskaya, Terzi, Sezener, Agrawal, de~Masson~d'Autume, Scholtes, Zaheer, Young, Gilsenan{-}McMahon, Austin, Blunsom, and Lazaridou}]{DBLP:conf/icml/LiskaKGTSAdSZYG22}
Adam Liska, Tom{\'{a}}s Kocisk{\'{y}}, Elena Gribovskaya, Tayfun Terzi, Eren Sezener, Devang Agrawal, Cyprien de~Masson~d'Autume, Tim Scholtes, Manzil Zaheer, Susannah Young, Ellen Gilsenan{-}McMahon, Sophia Austin, Phil Blunsom, and Angeliki Lazaridou. 2022.
\newblock \href {https://proceedings.mlr.press/v162/liska22a.html} {Streamingqa: {A} benchmark for adaptation to new knowledge over time in question answering models}.
\newblock In \emph{International Conference on Machine Learning, {ICML} 2022, 17-23 July 2022, Baltimore, Maryland, {USA}}, volume 162 of \emph{Proceedings of Machine Learning Research}, pages 13604--13622. {PMLR}.

\bibitem[{Longpre et~al.(2021)Longpre, Perisetla, Chen, Ramesh, DuBois, and Singh}]{DBLP:conf/emnlp/LongprePCRD021}
Shayne Longpre, Kartik Perisetla, Anthony Chen, Nikhil Ramesh, Chris DuBois, and Sameer Singh. 2021.
\newblock \href {https://doi.org/10.18653/v1/2021.emnlp-main.565} {Entity-based knowledge conflicts in question answering}.
\newblock In \emph{Proceedings of the 2021 Conference on Empirical Methods in Natural Language Processing, {EMNLP} 2021, Virtual Event / Punta Cana, Dominican Republic, 7-11 November, 2021}, pages 7052--7063. Association for Computational Linguistics.

\bibitem[{Mahapatra et~al.(2021)Mahapatra, Blagojevic, Bertorello, and Kumar}]{DBLP:journals/corr/abs-2112-13432}
Suchismit Mahapatra, Vladimir Blagojevic, Pablo Bertorello, and Prasanna Kumar. 2021.
\newblock \href {http://arxiv.org/abs/2112.13432} {New methods {\&} metrics for {LFQA} tasks}.
\newblock \emph{CoRR}, abs/2112.13432.

\bibitem[{Maynez et~al.(2020)Maynez, Narayan, Bohnet, and McDonald}]{DBLP:conf/acl/MaynezNBM20}
Joshua Maynez, Shashi Narayan, Bernd Bohnet, and Ryan~T. McDonald. 2020.
\newblock \href {https://doi.org/10.18653/v1/2020.acl-main.173} {On faithfulness and factuality in abstractive summarization}.
\newblock In \emph{Proceedings of the 58th Annual Meeting of the Association for Computational Linguistics, {ACL} 2020, Online, July 5-10, 2020}, pages 1906--1919. Association for Computational Linguistics.

\bibitem[{Min et~al.(2020)Min, Michael, Hajishirzi, and Zettlemoyer}]{DBLP:conf/emnlp/MinMHZ20}
Sewon Min, Julian Michael, Hannaneh Hajishirzi, and Luke Zettlemoyer. 2020.
\newblock \href {https://doi.org/10.18653/v1/2020.emnlp-main.466} {Ambigqa: Answering ambiguous open-domain questions}.
\newblock In \emph{Proceedings of the 2020 Conference on Empirical Methods in Natural Language Processing, {EMNLP} 2020, Online, November 16-20, 2020}, pages 5783--5797. Association for Computational Linguistics.

\bibitem[{M{\"{u}}ller et~al.(2020)M{\"{u}}ller, Rios, and Sennrich}]{DBLP:conf/amta/MullerRS20}
Mathias M{\"{u}}ller, Annette Rios, and Rico Sennrich. 2020.
\newblock \href {https://aclanthology.org/2020.amta-research.14/} {Domain robustness in neural machine translation}.
\newblock In \emph{Proceedings of the 14th Conference of the Association for Machine Translation in the Americas, {AMTA} 2020, Virtual, October 6-9, 2020}, pages 151--164. Association for Machine Translation in the Americas.

\bibitem[{Nema et~al.(2017)Nema, Khapra, Laha, and Ravindran}]{DBLP:conf/acl/NemaKLR17}
Preksha Nema, Mitesh~M. Khapra, Anirban Laha, and Balaraman Ravindran. 2017.
\newblock \href {https://doi.org/10.18653/v1/P17-1098} {Diversity driven attention model for query-based abstractive summarization}.
\newblock In \emph{Proceedings of the 55th Annual Meeting of the Association for Computational Linguistics, {ACL} 2017, Vancouver, Canada, July 30 - August 4, Volume 1: Long Papers}, pages 1063--1072. Association for Computational Linguistics.

\bibitem[{Pagnoni et~al.(2021)Pagnoni, Balachandran, and Tsvetkov}]{DBLP:conf/naacl/PagnoniBT21}
Artidoro Pagnoni, Vidhisha Balachandran, and Yulia Tsvetkov. 2021.
\newblock \href {https://doi.org/10.18653/v1/2021.naacl-main.383} {Understanding factuality in abstractive summarization with {FRANK:} {A} benchmark for factuality metrics}.
\newblock In \emph{Proceedings of the 2021 Conference of the North American Chapter of the Association for Computational Linguistics: Human Language Technologies, {NAACL-HLT} 2021, Online, June 6-11, 2021}, pages 4812--4829. Association for Computational Linguistics.

\bibitem[{Paszke et~al.(2019)Paszke, Gross, Massa, Lerer, Bradbury, Chanan, Killeen, Lin, Gimelshein, Antiga, Desmaison, K{\"{o}}pf, Yang, DeVito, Raison, Tejani, Chilamkurthy, Steiner, Fang, Bai, and Chintala}]{DBLP:conf/nips/PaszkeGMLBCKLGA19}
Adam Paszke, Sam Gross, Francisco Massa, Adam Lerer, James Bradbury, Gregory Chanan, Trevor Killeen, Zeming Lin, Natalia Gimelshein, Luca Antiga, Alban Desmaison, Andreas K{\"{o}}pf, Edward~Z. Yang, Zachary DeVito, Martin Raison, Alykhan Tejani, Sasank Chilamkurthy, Benoit Steiner, Lu~Fang, Junjie Bai, and Soumith Chintala. 2019.
\newblock \href {https://proceedings.neurips.cc/paper/2019/hash/bdbca288fee7f92f2bfa9f7012727740-Abstract.html} {Pytorch: An imperative style, high-performance deep learning library}.
\newblock In \emph{Advances in Neural Information Processing Systems 32: Annual Conference on Neural Information Processing Systems 2019, NeurIPS 2019, December 8-14, 2019, Vancouver, BC, Canada}, pages 8024--8035.

\bibitem[{Prabhumoye et~al.(2019)Prabhumoye, Quirk, and Galley}]{DBLP:conf/naacl/PrabhumoyeQG19}
Shrimai Prabhumoye, Chris Quirk, and Michel Galley. 2019.
\newblock \href {https://doi.org/10.18653/v1/n19-1269} {Towards content transfer through grounded text generation}.
\newblock In \emph{Proceedings of the 2019 Conference of the North American Chapter of the Association for Computational Linguistics: Human Language Technologies, {NAACL-HLT} 2019, Minneapolis, MN, USA, June 2-7, 2019, Volume 1 (Long and Short Papers)}, pages 2622--2632. Association for Computational Linguistics.

\bibitem[{Raffel et~al.(2020)Raffel, Shazeer, Roberts, Lee, Narang, Matena, Zhou, Li, and Liu}]{DBLP:journals/jmlr/RaffelSRLNMZLL20}
Colin Raffel, Noam Shazeer, Adam Roberts, Katherine Lee, Sharan Narang, Michael Matena, Yanqi Zhou, Wei Li, and Peter~J. Liu. 2020.
\newblock \href {http://jmlr.org/papers/v21/20-074.html} {Exploring the limits of transfer learning with a unified text-to-text transformer}.
\newblock \emph{J. Mach. Learn. Res.}, 21:140:1--140:67.

\bibitem[{Raunak et~al.(2021)Raunak, Menezes, and Junczys{-}Dowmunt}]{DBLP:conf/naacl/RaunakMJ21}
Vikas Raunak, Arul Menezes, and Marcin Junczys{-}Dowmunt. 2021.
\newblock \href {https://doi.org/10.18653/v1/2021.naacl-main.92} {The curious case of hallucinations in neural machine translation}.
\newblock In \emph{Proceedings of the 2021 Conference of the North American Chapter of the Association for Computational Linguistics: Human Language Technologies, {NAACL-HLT} 2021, Online, June 6-11, 2021}, pages 1172--1183. Association for Computational Linguistics.

\bibitem[{Su et~al.(2022)Su, Li, Zhang, Shang, Jiang, Liu, and Fung}]{DBLP:conf/acl/0003LZSJ0F22}
Dan Su, Xiaoguang Li, Jindi Zhang, Lifeng Shang, Xin Jiang, Qun Liu, and Pascale Fung. 2022.
\newblock \href {https://doi.org/10.18653/v1/2022.findings-acl.61} {Read before generate! faithful long form question answering with machine reading}.
\newblock In \emph{Findings of the Association for Computational Linguistics: {ACL} 2022, Dublin, Ireland, May 22-27, 2022}, pages 744--756. Association for Computational Linguistics.

\bibitem[{Tang et~al.(2022)Tang, Goyal, Fabbri, Laban, Xu, Yahvuz, Kryscinski, Rousseau, and Durrett}]{DBLP:journals/corr/abs-2205-12854}
Liyan Tang, Tanya Goyal, Alexander~R. Fabbri, Philippe Laban, Jiacheng Xu, Semih Yahvuz, Wojciech Kryscinski, Justin~F. Rousseau, and Greg Durrett. 2022.
\newblock \href {https://doi.org/10.48550/arXiv.2205.12854} {Understanding factual errors in summarization: Errors, summarizers, datasets, error detectors}.
\newblock \emph{CoRR}, abs/2205.12854.

\bibitem[{Wang et~al.(2022)Wang, Jatowt, and Yoshikawa}]{DBLP:conf/sigir/WangJY22}
Jiexin Wang, Adam Jatowt, and Masatoshi Yoshikawa. 2022.
\newblock \href {https://doi.org/10.1145/3477495.3531734} {Archivalqa: {A} large-scale benchmark dataset for open-domain question answering over historical news collections}.
\newblock In \emph{{SIGIR} '22: The 45th International {ACM} {SIGIR} Conference on Research and Development in Information Retrieval, Madrid, Spain, July 11 - 15, 2022}, pages 3025--3035. {ACM}.

\bibitem[{West et~al.(2022)West, Quirk, Galley, and Choi}]{DBLP:conf/acl/WestQGC22}
Peter West, Chris Quirk, Michel Galley, and Yejin Choi. 2022.
\newblock \href {https://doi.org/10.18653/v1/2022.findings-acl.294} {Probing factually grounded content transfer with factual ablation}.
\newblock In \emph{Findings of the Association for Computational Linguistics: {ACL} 2022, Dublin, Ireland, May 22-27, 2022}, pages 3732--3746. Association for Computational Linguistics.

\bibitem[{Wolf et~al.(2020)Wolf, Debut, Sanh, Chaumond, Delangue, Moi, Cistac, Rault, Louf, Funtowicz, Davison, Shleifer, von Platen, Ma, Jernite, Plu, Xu, Scao, Gugger, Drame, Lhoest, and Rush}]{DBLP:conf/emnlp/WolfDSCDMCRLFDS20}
Thomas Wolf, Lysandre Debut, Victor Sanh, Julien Chaumond, Clement Delangue, Anthony Moi, Pierric Cistac, Tim Rault, R{\'{e}}mi Louf, Morgan Funtowicz, Joe Davison, Sam Shleifer, Patrick von Platen, Clara Ma, Yacine Jernite, Julien Plu, Canwen Xu, Teven~Le Scao, Sylvain Gugger, Mariama Drame, Quentin Lhoest, and Alexander~M. Rush. 2020.
\newblock \href {https://doi.org/10.18653/v1/2020.emnlp-demos.6} {Transformers: State-of-the-art natural language processing}.
\newblock In \emph{Proceedings of the 2020 Conference on Empirical Methods in Natural Language Processing: System Demonstrations, {EMNLP} 2020 - Demos, Online, November 16-20, 2020}, pages 38--45. Association for Computational Linguistics.

\bibitem[{Wu and Hu(2018)}]{DBLP:conf/aaai/WuH18}
Yuxiang Wu and Baotian Hu. 2018.
\newblock \href {https://doi.org/10.1609/AAAI.V32I1.11987} {Learning to extract coherent summary via deep reinforcement learning}.
\newblock In \emph{Proceedings of the Thirty-Second {AAAI} Conference on Artificial Intelligence, (AAAI-18), the 30th innovative Applications of Artificial Intelligence (IAAI-18), and the 8th {AAAI} Symposium on Educational Advances in Artificial Intelligence (EAAI-18), New Orleans, Louisiana, USA, February 2-7, 2018}, pages 5602--5609. {AAAI} Press.

\bibitem[{Zhang and Choi(2021)}]{DBLP:conf/emnlp/ZhangC21}
Michael J.~Q. Zhang and Eunsol Choi. 2021.
\newblock \href {https://doi.org/10.18653/v1/2021.emnlp-main.586} {Situatedqa: Incorporating extra-linguistic contexts into {QA}}.
\newblock In \emph{Proceedings of the 2021 Conference on Empirical Methods in Natural Language Processing, {EMNLP} 2021, Virtual Event / Punta Cana, Dominican Republic, 7-11 November, 2021}, pages 7371--7387. Association for Computational Linguistics.

\bibitem[{Zhang et~al.(2020{\natexlab{a}})Zhang, Kishore, Wu, Weinberger, and Artzi}]{DBLP:conf/iclr/ZhangKWWA20}
Tianyi Zhang, Varsha Kishore, Felix Wu, Kilian~Q. Weinberger, and Yoav Artzi. 2020{\natexlab{a}}.
\newblock \href {https://openreview.net/forum?id=SkeHuCVFDr} {Bertscore: Evaluating text generation with {BERT}}.
\newblock In \emph{8th International Conference on Learning Representations, {ICLR} 2020, Addis Ababa, Ethiopia, April 26-30, 2020}. OpenReview.net.

\bibitem[{Zhang et~al.(2020{\natexlab{b}})Zhang, Merck, Tsai, Manning, and Langlotz}]{DBLP:conf/acl/ZhangMTML20}
Yuhao Zhang, Derek Merck, Emily~Bao Tsai, Christopher~D. Manning, and Curtis~P. Langlotz. 2020{\natexlab{b}}.
\newblock \href {https://doi.org/10.18653/v1/2020.acl-main.458} {Optimizing the factual correctness of a summary: {A} study of summarizing radiology reports}.
\newblock In \emph{Proceedings of the 58th Annual Meeting of the Association for Computational Linguistics, {ACL} 2020, Online, July 5-10, 2020}, pages 5108--5120. Association for Computational Linguistics.

\end{thebibliography}

\appendix
\label{sec:appendix}

\section{Benchmark Construction}
\label{sec:appendix:Benchmark_Construction}

Unlike previous work~\citep{DBLP:conf/emnlp/LongprePCRD021}, we follow the more natural setting where the transferred contextual knowledge is also factual. Besides, we make the question answerable as a necessary condition. Because we find the models prefer to generate hallucination when the contextual knowledge does not contribute to answering the question.

To construct long-form QA data, we reuse Debatepedia\citep{DBLP:conf/acl/NemaKLR17}, an abstractive summarization data, to supply our experiments. We choose this data due to its high abstractiveness and natural context transfer condition. We observe that there are lots of lexically similar examples, so we deduplicate examples whose Levenshtein distance is less than 4. 
This filtered dataset satisfies the format of $(q_i, c_i^+, a_i)$, and there are lots of questions paired with different contextual knowledge and answer. The examples with the same question are gathered, and one of them with the most distinctive answer is split into the development set. To enrich the contextual information of every case, we apply BM25 to retrieve negative knowledge $c_i^-$ from the whole dataset contexts via the question. Both relevant $c_i^+$ and irrelevant $c_i^-$ contexts are merged into $c_i$. Because if there is only $c_i^+$, the question $q_i$ is meaningless to position the positive context. In our basic setting, the contexts consist of 1 positive $c_i^+$ plus four negative $c_i^-$. The final processed dataset contains 2,549 training examples, 631 validation examples, and 598 test examples.

\section{Experimental Setting}
\label{sec:appendix:Experimental_Setting}

\begin{figure*}[t]
  \centering
  \includegraphics[width=0.93\textwidth]{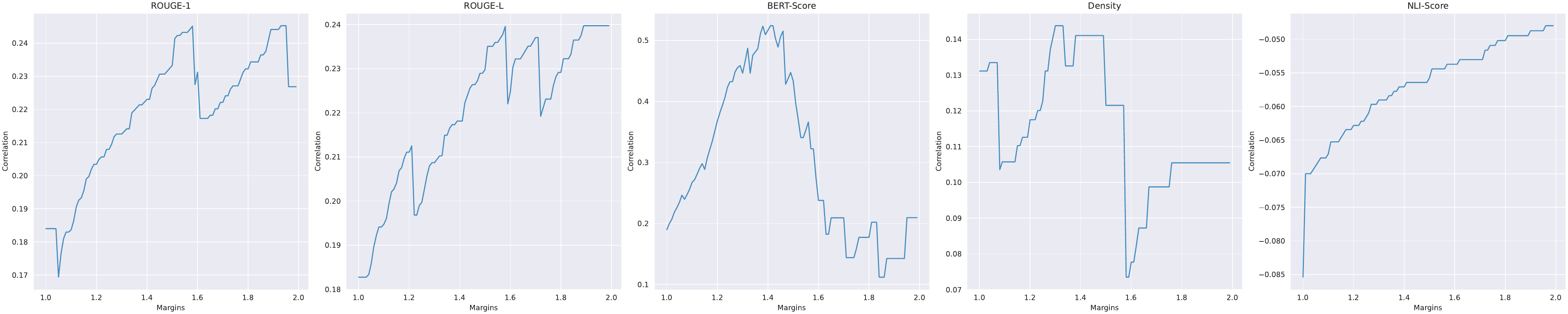}
  \caption{The Pearson correlation of margin failure ratio from basic metrics with different margins.}
  \label{correlation_under_margins}
\end{figure*}

\begin{table}[h]
\centering
\begin{tabular}{l|l}
\toprule
{\bfseries Parameter} & {\bfseries Value} \\
\hline
Learning Rate       &   $ 5 \times 10^{-5} $ \\
Batch Size          &   $ 16 $ \\
Accumulation Steps  &   $ 1 $ \\
Total Step          &   $ 4500 $ \\
Warmup Step         &   $ 150 $ \\
Evaluate Step       &   $ 150 $ \\
Weight Decay        &   $ 0.0 $  \\
Input Maximum Length       &   $ 512 $  \\
Output Maximum Length      &   $ 100 $  \\
Beam Size           &   $ 4 $ \\
\bottomrule
\end{tabular}
\caption{The experimental setting details. *Beam Size is the hyper-parameter of text generation in development and testing, while other parameters contribute to model training.}
\label{Tab:Settings}
\end{table}

We implement all the models using Pytorch~\citep{DBLP:conf/nips/PaszkeGMLBCKLGA19} and Transformers~\citep{DBLP:conf/emnlp/WolfDSCDMCRLFDS20} toolkit. The training and evaluation hyper-parameters are presented in \autoref{Tab:Settings}. We use Adam optimizer\citep{DBLP:journals/corr/KingmaB14} with the linear scheduler. All the training is started from the same random seed for a single round. We choose the best model by ROUGE-L score on the development set.

All the models are trained on a single NVIDIA V100 GPU with 32GB memory. Training BART-Large, BART-Large-xsum, FiD(BART-Large), FiD(BART-Large-xsum), T5-base, FiD(T5-base) takes approximately 3 hours. Training BART-base, FiD(BART-base), T5-small, FiD(T5-small) takes less than 1 hour.

\section{Meta Evaluation of MFR}
\label{sec:appendix:Meta_Evaluation}

We manually evaluate the grounding failure of context transfer on a small scale from test data in order that we can measure the Pearson Correlation between MFR and human labels. 
We ask two postgraduate students who major in natural language processing to manually evaluate the results. We also explain to them about memory hallucination under context transfer. We choose to label the generated results from FiD(BART-Large-xsum), as we observe this model hallucinates more than others. Human evaluation for more models is planned for future work.
We only label the examples whose generated answers get ROUGE-1 score of more than $40$ with the references in training data rather than all the examples in the test set. We believe only these cases could be hallucinated memory from training data. 
Notice that we only consider memory hallucination, which comes from training(fine-tuning phrase), while other hallucinations may also occur but are not taken into account. The final labelled data consist of $598$ items with only $22$ memory hallucination. Some case studies are presented in \autoref{Tab:human_evaluation_case_study}.

\begin{figure}[t]
  \centering
  \includegraphics[width=0.45\textwidth]{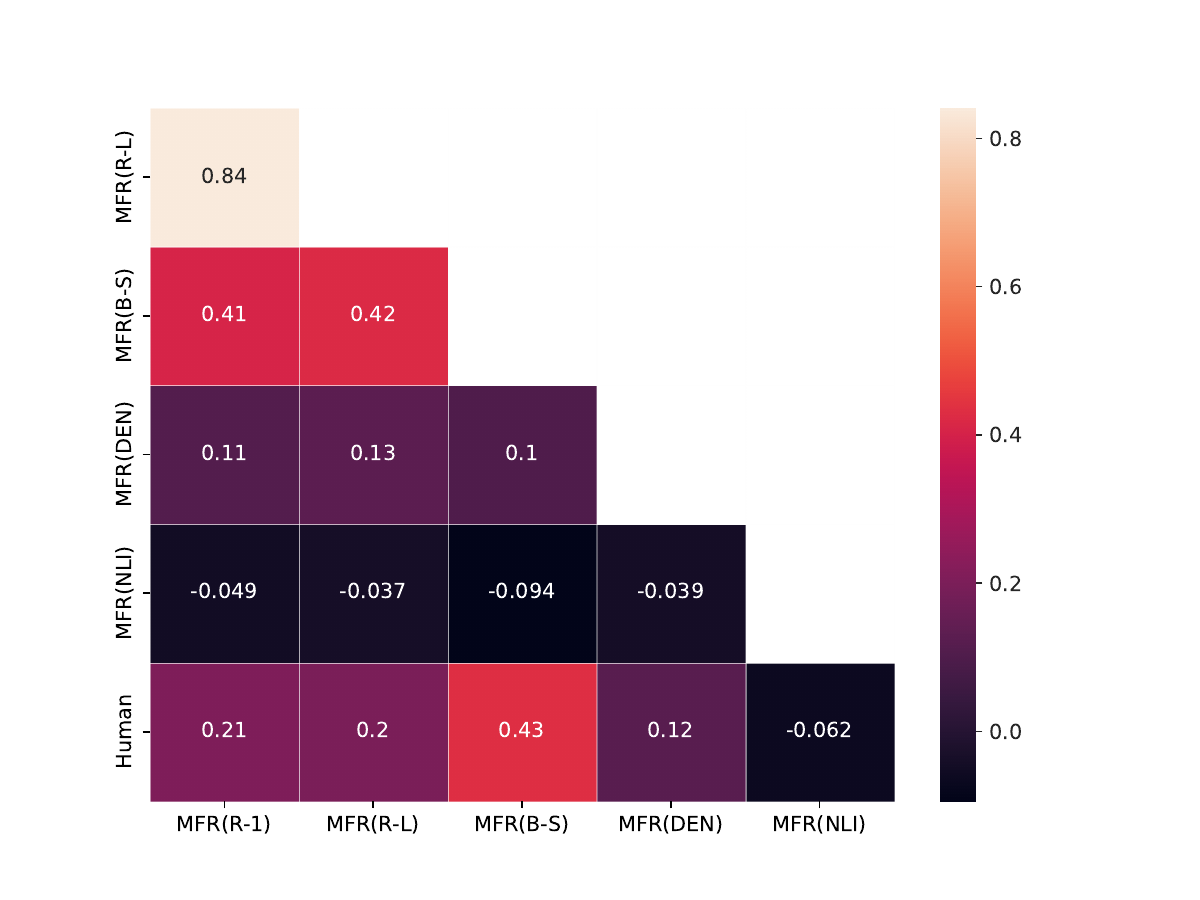}
  \caption{The Pearson correlation of margin failure ratio from each metric and human evaluation.}
  \label{correlation_under_metrics}
\end{figure}

We measure the Pearson correlation between different versions of MFR and human evaluation. We take the basic metrics $\Phi$ from two perspectives: the similarity with golden answers; the faithfulness to contextual knowledge. Concretely, for basic metrics of answer similarity, we use ROUGE(-1/L) and BERT-SCORE~\citep{DBLP:conf/iclr/ZhangKWWA20}; for basic metrics of knowledge faithfulness, we use Density\citep{DBLP:conf/naacl/GruskyNA18} and NLI-Score\footnote{We take the entailment probability from the RoBERTa-Large classifier fine-tuned on MNLI as NLI-Score.}.  As depicted in \autoref{correlation_under_metrics}, all automatic metrics are only a little related to each other, except MFR(ROUGE-1) and MFR(ROUGE-L). There is even little relationship between MFR(NLI-Score) and human evaluation. MFR(BERT-Score) performs best correlatively with human evaluation, so we take MFR(BERT-Score) as the main measure in this work.

We also measure the influence of the margin $m$. For each metric $\Phi$ in MFR, we experiment with its margin varying from $1.00$ to $2.00$ with a stripe of $0.01$. As shown in \autoref{correlation_under_margins}, the margin $m$ has a great impact on the human correlation of MFR and different basic metrics achieve the best performance at different margins. Although the intuitively chosen margin $m=1.25$ is not the perfect hyperparameter of BERT-Score, it still has a relatively strong correlation with Pearson Correlation of $0.43$.

\begin{table*}[t]
\scriptsize
\centering
\begin{tabular}{l|l|l|l}

\toprule

{\bfseries Testing Data} & {\bfseries Training Data} & {\bfseries R-L} & {\bfseries Label} \\
\hline

\makecell[l{m{6.1cm}}]{
QUESTION: \\
genocide ? can the violence in darfur be considered genocide ? \\ \\
CONTEXT: \\
joschka fischer . former german foreign minister and vice chancellor from 1998 to 2005 . `` the eu must act in darfur . targeted sanctions would be a real step towards stopping the killing . '' april 19th 2007 - `` ... there insufficient political will for an international force [ in darfur ] ... '' \\ \\
GOLDEN ANSWER: \\
there is insufficient political will for military intervention in darfur \\ \\
PREDICTED ANSWER: \\
the violence in darfur could be considered genocide. \\ \\
} 
&
\makecell[l{m{6.1cm}}]{
QUESTION: \\
genocide ? can the violence in darfur be considered genocide ? \\ \\
CONTEXT: \\
genocide is defined by most to include the systematic murders of a group of peoples as well as deliberate displacement and abuse . more than \# \# people have died since \# with other estimates ranging up to \# \# according to amnesty international and the un . over \# million people have become displaced and many are in danger of starvation due to lack of water and food . conclusively darfur is the worst humanitarian abuse in africa . to the extent that the janjaweed is systematically overseeing this mass-murder and to the extent that the government is involved in supporting the janjaweed darfur 's crisis can be considered a genocide . \\ \\
GOLDEN ANSWER: \\
the violence in darfur could be considered genocide
} 
&
$22.22/100.00$
&
True
\\
\hline

\makecell[l{m{6.1cm}}]{
QUESTION: \\
changing menus : will mandatory calorie counts compel restaurants to improve menus ? \\ \\
CONTEXT: \\
restaurants that get caught under-reporting calories on their menus may face not only fines from the government but also significant pr problems as stories of their manipulations reach and turn-off their customers . \\ \\
GOLDEN ANSWER: \\
restaurants will not under-report calories and risk pr backlash . \\ \\
PREDICTED ANSWER: \\
restaurants under-report calories on menus  \\
} 
&
\makecell[l{m{6.1cm}}]{
QUESTION: \\
changing menus : will mandatory calorie counts compel restaurants to improve menus ? \\ \\
CONTEXT: \\
`` calorie disclosures fail to weigh whole enchilada '' . wall street journal . july 8 2009 : `` scripps television stations sent several menu items to testing labs and found some big deviations from posted calorie content most of them making menu items appear healthier than they are . for example two tests of applebee 's cajun-lime tilapia meal found about 400 calories compared with the posted total of 310 . '' this means that restaurants may simply choose to lower their reporting of calories instead of actually lower the calories in the foods they are serving . \\ \\
GOLDEN ANSWER: \\
restaurants frequently under-report calories on menus
} 
&
$42.86/90.91$
&
False 
\\
\hline

\makecell[l{m{6.1cm}}]{
QUESTION: \\
wealthy : is a progressive tax system fair to the wealthy ? \\ \\
CONTEXT: \\
david n. mayer . `` wealthy americans deserve real tax relief on principle '' . ashbrook center . october \# - `` there is no correlation between the amount of taxes an american pays and whatever benefits if any he receives ; indeed a wealthy person may get fewer government services than a poorer person . '' \\ \\
GOLDEN ANSWER: \\
the rich do not necessarily benefit more from taxes/system \\ \\
PREDICTED ANSWER: \\
progressive tax system unfairly benefits the wealthy 
} 
&
\makecell[l{m{6.1cm}}]{
QUESTION: \\
wealthy : is a progressive tax system fair to the wealthy ? \\ \\
CONTEXT: \\
it is unfair that people who earn more should pay at a progressive rate . even on a standard rate they already pay more tax because they have a higher taxable income . therefore progressive tax rates are a form of double taxation as higher earners pay tax on more income and then at a high level . this is further unfair to them since high earners are the least likely group to benefit from much taxpayer-funded activity e.g . welfare . \\ \\
GOLDEN ANSWER: \\
flat tax fairly has wealthy pay proportionally more in taxes .
} 
&
$12.50/23.53$
&
True
\\
\hline

\makecell[l{m{6.1cm}}]{
QUESTION: \\
militia : does the \# nd amendment secure an individual right to form an independent militia ? \\ \\
CONTEXT: \\
an armed citizenry empowers citizens to protect themselves so that a big government does n't have to . \\ \\
GOLDEN ANSWER: \\
in order to form a militia citizens require guns and a right to own them \\ \\
PREDICTED ANSWER: \\
the \# nd amendment secured an individual right to bear arm for the purpose of self-defense
} 
&
\makecell[l{m{6.1cm}}]{
QUESTION: \\
militia : does the \# nd amendment secure an individual right to form an independent militia ? \\ \\
CONTEXT: \\
an armed citizen can places a checking on inappropriate cops power and the emergence of a cops state . \\ \\
GOLDEN ANSWER: \\
\# nd amendment secured equally the right of the militia and the individual to arms .
} 
&
$14.29/42.86$
&
False
\\

\bottomrule
\end{tabular}
\caption{Case study of human evaluation. The $X/Y$ in R-L denotes the ROUGE-L score of predicted answer with the golden answer in testing($X$) or training($Y$) data. And Label denotes the human label for memory hallucination under knowledge transfer.}
\label{Tab:human_evaluation_case_study}
\end{table*}





\end{document}